%% file: main.tex
\documentclass[a4paper]{cas-dc}

\usepackage[authoryear,longnamesfirst]{natbib}

\usepackage{float}
\usepackage{amsmath}
\usepackage{multirow}
\usepackage{makecell}
\usepackage{float}
\usepackage{tabularray}
\usepackage{array,ragged2e}
\usepackage{textcomp}
\usepackage{listings}
\usepackage{booktabs}
\usepackage{xcolor}
\usepackage{subcaption}
\definecolor{darkred}{RGB}{139,0,0}
\definecolor{codebg}{gray}{0.95}

\begin{document}
\let\WriteBookmarks\relax
\def\floatpagepagefraction{1}
\def\textpagefraction{.001}

\shorttitle{Data-efficient domain-aware flood depth prediction}
\shortauthors{Huang et al.}

\title[mode=title]{Data-efficient flood depth prediction through domain-aware coreset selection and tabular foundation models}

\author[1]{Lipai Huang}[orcid=0009-0005-3732-7319]
\cormark[1]
\ead{lipai.huang@tamu.edu}
\credit{Conceptualization, Methodology, Software, Writing - original draft}

\author[2]{Adithi Srinath}
\credit{Software, Validation}

\author[2]{Manas Singh}
\credit{Software, Validation}

\author[1,3]{Junwei Ma}
\credit{Methodology, Writing - review and editing}

\author[1,4]{Ali Mostafavi}
\credit{Supervision, Writing - review and editing}

\affiliation[1]{organization={Urban Resilience.AI Lab, Zachry Department of Civil and Environmental Engineering, Texas A\&M University},
    city={College Station},
    state={TX},
    country={USA}}

\affiliation[2]{organization={Department of Computer Science and Engineering, Texas A\&M University},
    city={College Station},
    state={TX},
    country={USA}}

\affiliation[3]{organization={Resilitix Intelligence LLC},
    city={Houston},
    state={TX},
    country={USA}}
    
\affiliation[4]{organization={Institute for a Disaster Resilient Texas, Texas A\&M University},
    city={College Station},
    state={TX},
    country={USA}}

\cortext[1]{Corresponding author}

\input{0_abstract}

\begin{highlights}
\item A two-stage coreset jointly stratifies by storm return period and spatial structure.
\item A vanilla foundation model matches the watershed-level baseline with a 50k coreset.
\item In-context learning transfers to held-out watersheds within one metropolitan region.
\item A coreset foundation model beats the full-data reference on a far-OOD storm.
\end{highlights}

\begin{keywords}
Domain-aware coreset construction \sep In-context learning \sep Tabular foundation model \sep Flood depth prediction \sep Cross-watershed transferability \sep Hydrodynamic surrogate
\end{keywords}

\maketitle

\input{1_INTRODUCTION}
\input{2_RELATED_WORK}
\input{3_DATA}

\input{4_METHODOLOGY}
\input{5_EXPERIMENTS_AND_RESULTS}
\input{6_CONCLUSION}

\section*{Code Availability}
The code that supports the findings of this study is available from the corresponding author upon request.

\section*{Data Availability}
The data that support the findings of this study is available from the corresponding author upon request.
\balance
% \printcredits   % uncomment to print CRediT contribution table

\bibliographystyle{cas-model2-names}
\bibliography{ref}

\end{document}

%% file: 0_abstract.tex
\begin{abstract}
Near-real-time flood depth prediction demands surrogate models that are accurate, fast, and transferable across watersheds.
Supervised surrogates can match physics-based simulators in accuracy but need millions of training rows per watershed and cannot extrapolate beyond their original mesh.
We propose a domain-aware coreset construction pipeline that conditions a tabular foundation model at inference time.
The pipeline stratifies storms by return period and most-affected watershed, then samples hexagons with a target-aware spatial selector.
With $0.7\%$ of the per-watershed training pool, the model attains a mean $R^2$ of $0.663$ across nine Houston-area watersheds, statistically indistinguishable from the supervised reference ($R^2 = 0.673$).
It transfers to held-out watersheds within the same metropolitan region without task-specific retraining, staying ahead of a coreset-trained supervised baseline.
On real storms it exceeds the supervised reference on a far out-of-distribution case and trails it on a mixed case that is only moderately out of distribution.
Domain-aware coreset construction lets tabular foundation models deliver data-efficient flood predictions that transfer across watersheds within a region without per-watershed training.
\end{abstract}
%%%%%%

%% file: 1_INTRODUCTION.tex
\section{Introduction}\label{sec:1}

\begin{figure*}[t]
    \centering
    \includegraphics[width=\textwidth]{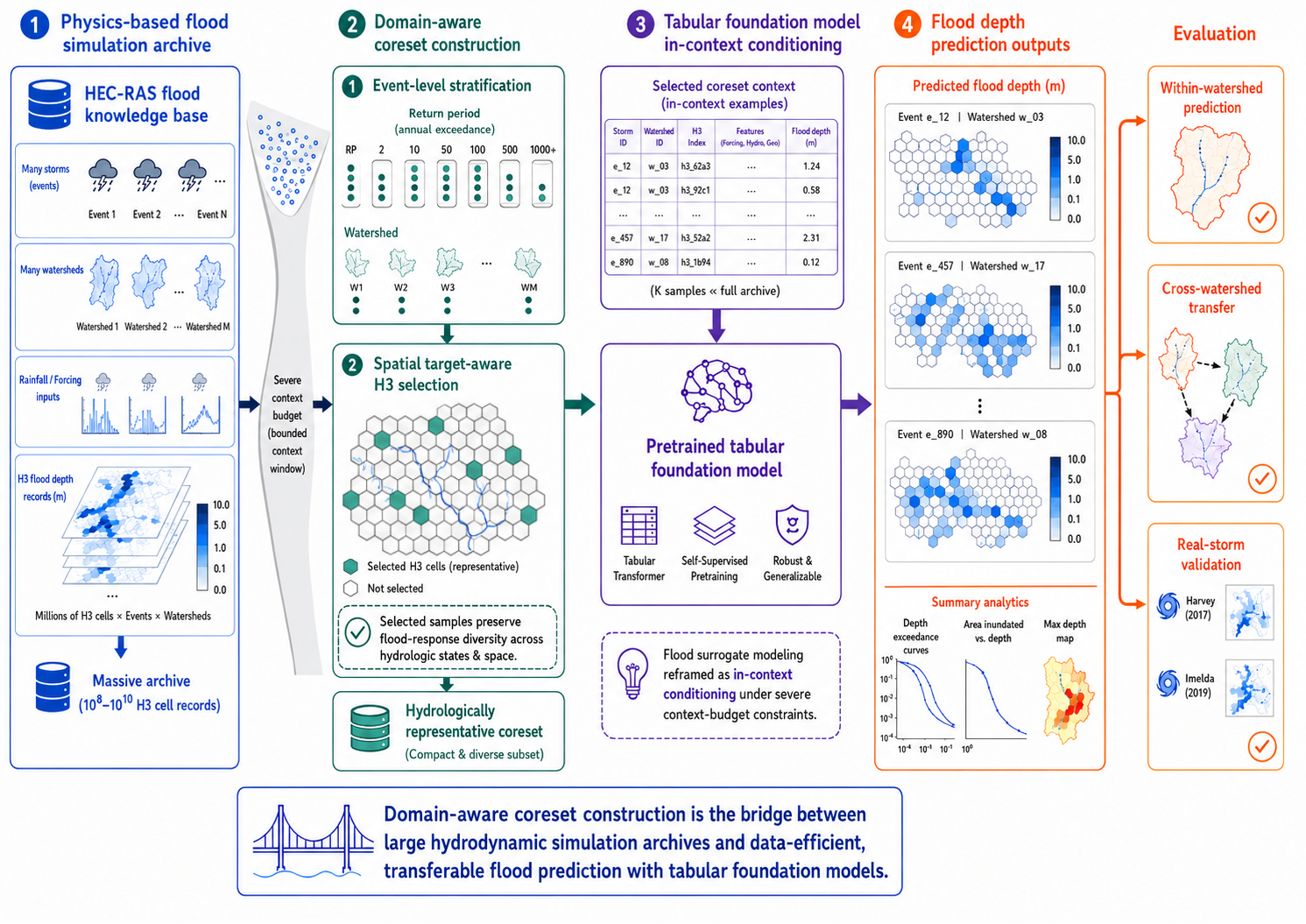}
    \caption{Conceptual overview of the proposed approach. (1)~Physics-based flood simulation archive: a HEC-RAS knowledge base of 592 synthetic storm events across nine Houston-area watersheds, on the order of $10^8$ event-hexagon rows, with storm metadata, watershed boundaries, and NOAA Atlas~14 return-period labels. (2)~Domain-aware coreset construction: a two-stage pipeline compresses the archive into a compact, hydrologically representative subset. Stage~1 stratifies events by return period and most-affected watershed, and Stage~2 selects H3 Level~10 hexagons with a target-aware facility-location strategy. (3)~Tabular foundation model in-context conditioning: the coreset conditions a pretrained tabular foundation model at inference, with no per-watershed fine-tuning. (4)~Flood-depth prediction outputs: the model returns peak inundation depth for query hexagons. The pipeline is evaluated under three protocols: within-watershed accuracy, cross-watershed leave-one-out transfer, and real-storm validation on Hurricane Harvey and Tropical Storm Imelda. (The schematic was prepared with the assistance of ChatGPT-5.5.)}
    \label{fig:main_flow}
\end{figure*}

Timely flood depth information supports emergency managers and infrastructure operators during extreme weather events, and also underpins downstream assessments of flood exposure \citep{yin2023unsupervised}, mobility disruption, and community resilience \citep{yin2026deep}.
Yet generating it at scale remains computationally demanding \citep{li2025parametric}.
Physics-based hydrodynamic models such as the Hydrologic Engineering Center's River Analysis System (HEC-RAS) provide the engineering reference for inundation depth prediction.
Their cost grows quickly with simulated extent and mesh resolution, especially when many storm events must be screened.
A direct HEC-RAS sweep is therefore impractical for near-real-time forecasting or broad scenario testing \citep{ma2026uncovering}.
Machine learning (ML) surrogates trained on simulator output have become a standard alternative and offer near-instant inference once trained \citep{bentivoglio2022deep, mosavi2018flood}.

Existing surrogate work falls into two patterns.
The first trains one model per mesh cell, capturing local dynamics at fine resolution while producing a fragmented collection that cannot be applied beyond the original mesh \citep{lee2024predicting}.
The second trains a single supervised model on the full knowledge base of one watershed, which yields strong in-watershed accuracy but binds the predictor to its training region \citep{zahura2020training}.
Both patterns share the same operational drawback: each requires millions of training rows, and adding a new watershed or recalibrating after an event-regime shift entails another full training pass.
Tree-based gradient boosting \citep{chen2016xgboost} still dominates this space because deep tabular methods continue to underperform tree ensembles on structured input \citep{grinsztajn2022why}.
Deep learning has produced striking results on adjacent hydrologic tasks such as rainfall-runoff modeling \citep{kratzert2018rainfall}, but for spatially distributed flood-depth surrogates the tree-based pattern remains the practical default.

Transformer and foundation-model architectures have recently been adapted to structured disaster-management tasks, including post-disaster building-damage categorization~\citep{xiao2025damagecat}, multimodal impact assessment~\citep{xiao2026crisisense}, and graph-based damage prediction~\citep{esparza2026graph}.
Tabular foundation models (TFMs) such as TabPFN \citep{hollmann2025accurate} and TabICL \citep{qu2025tabicl} offer a different path.
A TFM is a single transformer pretrained on a wide distribution of synthetic tabular tasks.
It solves a new task at inference time by conditioning on a labeled context set rather than through gradient updates.
This in-context learning (ICL) framing avoids per-task retraining, and a sufficiently strong pretrained backbone can rival task-specific supervised models on tabular benchmarks.
However, the context windows of the tabular foundation models considered in this work are bounded at roughly $10^4$ to $10^5$ rows \citep{hollmann2025accurate, grinsztajn2025tabpfn, qu2025tabicl}, while a single watershed's simulation knowledge base typically reaches millions.
Choosing which rows to put into the context becomes the central operational question.
Naive options such as random sampling or feature-only facility location ignore the spatial autocorrelation and event-magnitude imbalance that govern flood data.
Building a high-quality coreset for a TFM in this setting therefore depends on encoding hydrologic and geographic prior knowledge into the selection step.

Figure~\ref{fig:main_flow} summarizes the proposed approach.
We propose that a TFM conditioned on a carefully built coreset is a feasible flood-depth surrogate at the watershed level.
We then test how far the same configuration transfers to held-out watersheds under a leak-free protocol.
To support both questions we develop a domain-aware coreset construction pipeline that operates in two stages.
Stage one stratifies storm events on two axes: storm return period (RP) and most-affected watershed.
The dual stratification ensures that rare high-RP events and their host watersheds survive sampling noise.
Stage two samples from a hierarchical hexagonal (H3) grid at Level 10 with cell edge length of approximately 75~m.
The selector combines facility-location coverage in static-feature space with a z-scored target-depth signal. For cross-watershed evaluation we adopt a leave-one-out (LOO) protocol in which the held-out target watershed contributes neither training rows nor context.

The contributions of this paper are:
\begin{itemize}
    \item A two-stage domain-aware coreset construction pipeline that combines event-level stratification by return period and watershed with target-aware hexagon selection, encoding hydrologic and geographic priors in the context.
    \item A demonstration that a vanilla TFM conditioned on the domain-aware coreset (about $0.7\%$ of the per-watershed training pool) reaches an average $R^2$ across nine Houston-area watersheds that is statistically indistinguishable from a watershed-level supervised reference trained on the full knowledge base.
    \item A leak-free leave-one-out protocol with two source-selection modes (neighboring versus all other watersheds), showing that the same vanilla TFM transfers to held-out watersheds within the same metropolitan region without retraining and outperforms a coreset-trained supervised baseline at most context sizes in both modes.
\end{itemize}

%% file: 2_RELATED_WORK.tex
\section{Related Work}\label{sec:2}

\subsection{Return Period for Storm Stratification}
Hydrology distinguishes pluvial return period (RP) from fluvial RP by their driver. Pluvial RP ranks events by short-duration rainfall intensity and supports urban drainage design where local rainfall dominates. Fluvial RP ranks events by streamflow magnitude at a gauge and supports riverine flood mapping where upstream runoff dominates. The two can rank the same storm differently, so combined RP frameworks treat both drivers jointly and are preferred in mixed-regime basins where rainfall and runoff co-vary \citep{zscheischler2018future, wahl2015compound}. All variants share the same statistical foundation: a $T$-year storm has a $1/T$ chance of being equaled or exceeded in any single year. RP underpins floodplain mapping, hydraulic structure design, and insurance rate setting. In the United States the reference curves are published in NOAA Atlas 14 \citep{noaaatlas14}, which maps storm duration and accumulated rainfall depth to RP at the county level. For supervised flood-prediction datasets, synthetic storm libraries tend to oversample moderate events, so intentional stratification by RP is needed to keep rare, high-impact storms from being under-represented in training \citep{bentivoglio2022deep}.

\subsection{Coreset Selection}
A coreset is a small, weighted subset of a dataset chosen so that a model trained or conditioned on the subset behaves similarly to one trained on the full data \citep{phillips2017coresets, mirzasoleiman2020coresets, bachem2017practical}. Coreset selection methods fall into three broad families. Uniform random sampling provides an unbiased default but covers feature space inefficiently when data are imbalanced. Geometric methods such as facility location \citep{lin2011submodular, wei2015submodularity} and core-set covering \citep{sener2018active} greedily pick samples that maximize coverage of a feature-space kernel, offering diversity guarantees but ignoring labels. Target-aware methods incorporate label statistics: gradient-based selectors \citep{killamsetty2021glister, killamsetty2021gradmatch} pick samples whose gradients best approximate the full-batch update, proxy-based selectors \citep{coleman2020selection} rank candidates with a cheaper surrogate model, and pruning metrics such as forgetting score or supervised classification margin separate redundant from informative examples \citep{sorscher2022beyond}. In the foundation-model era, coreset selection has taken on a second role of choosing in-context examples that condition a pretrained predictor at inference time \citep{hollmann2025accurate, thomas2024retrieval}. Most of these methods assume samples are independent, an assumption that breaks in geophysical applications where features are strongly autocorrelated, causing naive feature-space selection to cluster samples geographically and leave parts of the domain unrepresented \citep{roberts2017cv, meyer2018improving}.

\subsection{In-Context Learning for Tabular Prediction}
Tabular foundation models (TFMs) cast tabular prediction as in-context learning: a transformer is pretrained once on a large distribution of synthetic tasks and, at inference, conditions on a labeled context set $(\mathbf{X}_{ctx}, \mathbf{y}_{ctx})$ to predict labels for a query set $\mathbf{X}_q$ without gradient updates. The framing originates in the prior-data fitted networks of \citet{muller2022transformers} and was specialized to tabular classification and regression by TabPFN \citep{hollmann2023tabpfn, hollmann2025accurate}. Subsequent releases progressively expand the supported context: TabPFN-v2.5 \citep{grinsztajn2025tabpfn} reaches roughly $5 \times 10^4$ rows, and TabPFN-v2.6 \citep{priorlabs2025tabpfn26} extends this to $10^5$ rows. TabICL \citep{qu2025tabicl} targets even larger context sizes through a column-then-row attention mechanism. A separate strand augments these backbones with task-specific fine-tuning \citep{thomas2024retrieval}, with full fine-tuning recently shown to be a stable baseline for TabPFN-v2 \citep{rubachev2025finetuning}. Transfer learning has likewise been used to improve tabular prediction from limited engineering data~\citep{pak2023knowledge}.

\subsection{Out-of-Distribution Evaluation}
Spatially distributed predictors are vulnerable to two distinct evaluation failures. Standard random $k$-fold cross-validation underestimates prediction error on spatially autocorrelated data, because train and test folds remain close in feature and physical space \citep{roberts2017cv}. Spatial cross-validation schemes such as Leave-Location-Out hold out entire spatial blocks to break that contamination \citep{meyer2018improving}. A separate question is whether predictions made outside the training distribution can be trusted at all: \citet{meyer2021predicting} formalize this as the Area of Applicability of a spatial model, and the WILDS benchmark \citep{koh2021wilds} catalogues representative distribution shifts in machine learning.

%% file: 3_DATA.tex
\section{Data}\label{sec:3}

\begin{figure}[pos=!h]
    \centering
    \includegraphics[width=\linewidth]{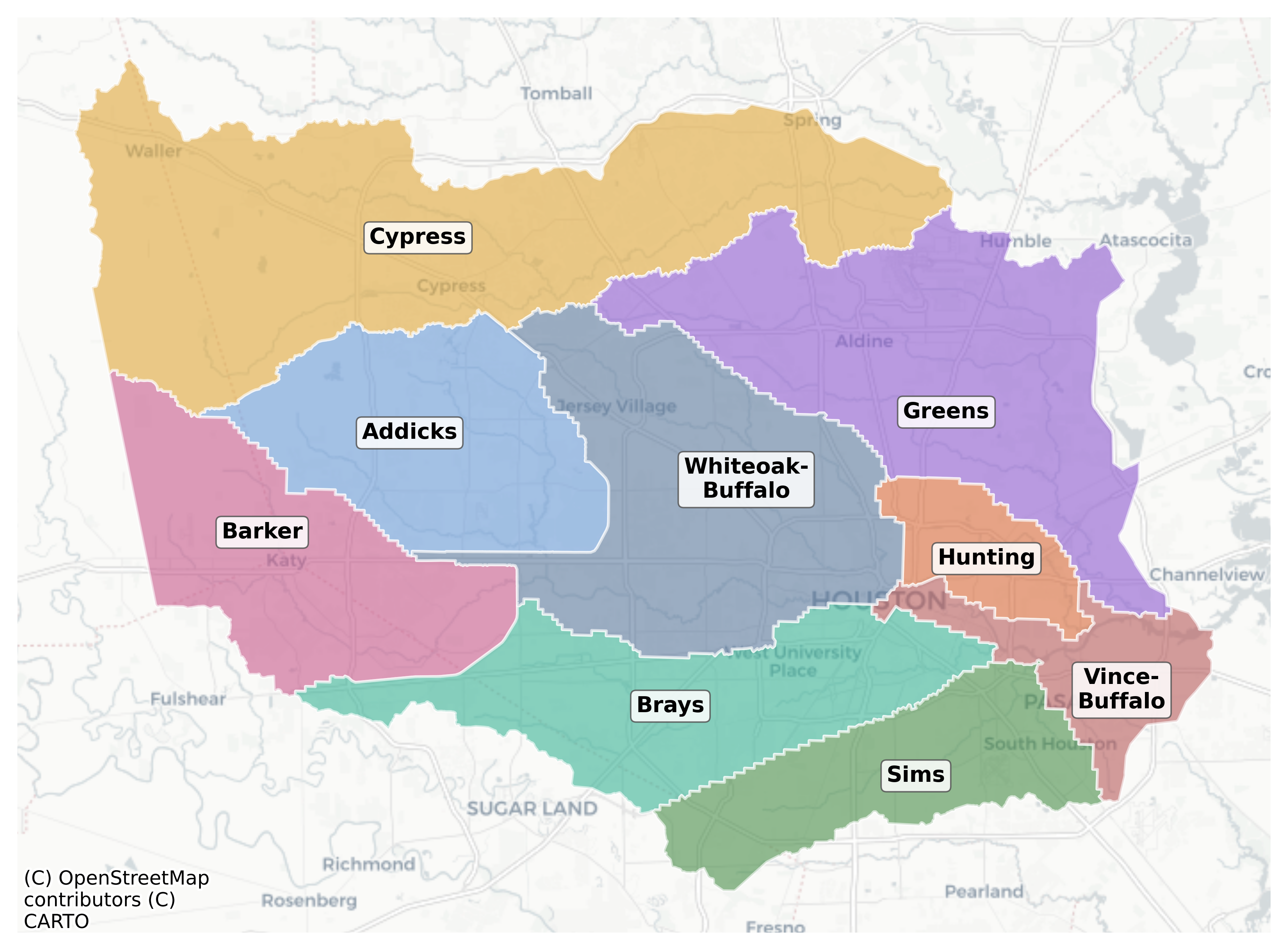}
    \caption{The nine Houston-area watersheds, dissolved from the HEC-RAS simulation mesh. Basemap rendered with contextily (\url{https://contextily.readthedocs.io/en/latest/index.html}) using OpenStreetMap and CARTO tiles.}
    \label{fig:watersheds}
\end{figure}

The primary data is the MaxFloodCast HEC-RAS simulation database \citep{lee2024predicting} covering nine watersheds in Harris County, Texas (Figure~\ref{fig:watersheds}).
Harris County is the largest county in the Greater Houston Metropolitan Statistical Area, with a substantially flat topography ranging from roughly $-12$~m to $91$~m above mean sea level and a population exceeding $4.5$ million.
Two main hydrologic systems organize the county: Cypress Creek in the north and the Buffalo Bayou system across the central and southern portions, both draining eastward through the San Jacinto River and the Ship Channel into the Gulf of Mexico.

The database contains $592$ synthetic storm events generated by applying a Rasterized Time-series Resampling Method to historic storms in the area, with durations from $1$ to $33$ hours and hourly rainfall grids at approximately $1{,}010$~m resolution.
Flood inundation depths are simulated by HEC-RAS 2D over an unstructured mesh of $26{,}301$ cells (nominal size $\approx 366$~m), refined along major watercourses and breaklined at high-elevation features.
From this database we inherit the cell-level peak inundation depth target and the per-event rainfall metadata.
All other inputs are processed independently in this study.

To establish a uniform spatial unit across all variables, we discretize each mesh cell into H3 Level 10 hexagons (about $75$~m edge length) via polygon polyfill ($\approx 6.7$ hexagons per mesh cell on average).
We use H3 \citep{uber2018h3} mainly for its indexing, since each hexagon has a unique identifier that turns neighbor lookup, aggregation, and joins across data sources into identifier operations rather than geometry computation.
Level 10 is the resolution at which our group maintains its static feature library for the region.
It is finer than the mesh and coarser than the source rasters, so it preserves sub-mesh variation in the static features without inflating the row count further.
We did not tune it for this study.
Static geophysical features are extracted from external geospatial datasets and resampled onto the resulting $177{,}330$ hexagons.
The HEC-RAS depth output and event-level rainfall metadata are likewise resampled to the same H3 Level 10 grid.
Hexagons in the same mesh cell share that cell's dynamic features (rainfall, depth) but carry their own static features, yielding $592$ events $\times$ $177{,}330$ hexagons $\approx 105$ million rows with no missing values.
Section~\ref{sec:5-exp1-perws} quantifies what this label sharing means for coreset coverage.
Per-watershed counts range from $2.2$~M (Hunting Bayou) to $14.1$~M (Cypress Creek), with a mean of about $6.9$~M.

\begin{table*}[h]
\centering
\caption{Key variables used in modeling and event stratification, with data sources.}
\label{tab:variables}
\begin{tabular}{@{}lll@{}}
\toprule
Category & Variable & Data Source \\
\midrule
\multirow{5}{*}{Topography}
  & Elevation & National Elevation Dataset (NED) \\
  & Imperviousness & National Land Cover Database (NLCD) \\
  & Topographic Wetness Index (TWI) & NED + NLCD \\
  & Road Density within 500\,m radius & TxDOT roadway network \\
  & Distance to Coast & National Hydrography Dataset Plus High Resolution (NHDPlusHR) \\
\midrule
\multirow{3}{*}{Hydrology}
  & Height above Nearest Drainage (HAND) & NOAA National Water Model \\
  & Distance to Nearest Stream & NHDPlusHR \\
  & Distance to Stream (order $\geq 4$) & NHDPlusHR \\
\midrule
\multirow{3}{*}{\makecell{Aggregated\\Event-specific}}
  & Cumulative Rainfall & HEC-RAS 2D \\
  & Peak Rainfall Intensity & HEC-RAS 2D \\
  & Precipitation Duration & HEC-RAS 2D \\
\midrule
Event Annotation & Return Period (RP) & NOAA Atlas 14 \\
\bottomrule
\end{tabular}
\end{table*}

Each row carries 11 predictors and a target (Table~\ref{tab:variables}).
Eight static predictors describe local topography and hydrology: elevation, imperviousness, Topographic Wetness Index (TWI), road density within a 500\,m radius, distance to coast, Height above Nearest Drainage (HAND), distance to the nearest stream, and distance to the nearest stream of order $\geq 4$.
Three dynamic predictors describe each storm: cumulative rainfall, peak rainfall intensity, and precipitation duration.
Among the static predictors, TWI follows the standard $\ln(\mathrm{SCA}/\tan\beta)$ formulation under $D_\infty$ flow routing \citep{tarboton1997new} computed in SAGA GIS \citep{conrad2015system}, and road density is the fraction of road area within a 500\,m circular kernel applied to a 15\,m-buffered roadway raster.
The target is the per-event peak inundation depth at each hexagon simulated by HEC-RAS in feet.
Rainfall descriptors are in inches and inches per hour.
The model predicts one peak-depth value per event and hexagon from static geospatial features and aggregated rainfall descriptors, not a time-evolving nowcast that updates depth as new rainfall or gauge data arrive.

Each storm is annotated with two derived fields used for stratification.
The most-affected watershed is the watershed whose cells have the highest mean inundation depth during that storm.
The RP bin assigns each storm to the largest exceeded NOAA Atlas 14 \citep{noaaatlas14} county-median threshold at the PFDS duration class matching the event's storm duration, with the procedure detailed in Section~\ref{sec:4-rp}.
RP is used as a stratification label rather than a model input.
Bin populations cover the range from below 1-year through 1{,}000-year, with $42$ events each in the $500$-year and $1{,}000$-year bins.

The $592$ events are split into $350$ for training, $121$ for validation, and $121$ for testing under double stratification on RP bin and most-affected watershed, with a post-processing pass that prevents any single watershed from being underrepresented at test time.
Within the test set, $42$ stratified-subsample events serve as the query set in all experiments.

For external validation we use two real storms simulated with observed rainfall: Hurricane Harvey (August 2017) and Tropical Storm Imelda (September 2019).
Rainfall inputs come from Harris County Flood Control District rain-gage observations \citep{hcfcd2023fws} interpolated through Thiessen polygons, while depth targets are produced by HEC-RAS 2D under the same setup as the synthetic events and resampled onto the same H3 L10 grid.
Because the real events share the mesh and the H3 grid, every hexagon carries the same static features as in the synthetic knowledge base, so any difference between a real storm and the synthetic training range lies in the three rainfall variables.
A query cell is in-distribution when its cumulative rainfall, peak rainfall intensity, and duration all fall within the 5th to 95th percentile range of the watershed's $50\text{k}$-row random pool, and out-of-distribution (OOD) otherwise.
Hurricane Harvey lies entirely outside this envelope and is treated as fully OOD, while Tropical Storm Imelda mixes in-distribution and OOD cells.

%% file: 4_METHODOLOGY.tex
\section{Methodology}\label{sec:4}

\subsection{Return Period Construction}\label{sec:4-rp}

Each storm event is labeled with a single RP bin used to stratify the dataset split and the Stage 1 event selection.
The event is first assigned a NOAA Atlas 14 duration class $d^\ast$ based on its storm duration $d_e$:
\begin{equation}\label{eq:rp_duration_class}
d^\ast = \begin{cases} 6\,\mathrm{h}, & d_e \leq 6\,\mathrm{h} \\ 12\,\mathrm{h}, & 6\,\mathrm{h} < d_e \leq 12\,\mathrm{h} \\ 24\,\mathrm{h}, & d_e > 12\,\mathrm{h} \end{cases}
\end{equation}
which selects the smallest NOAA Atlas 14 \citep{noaaatlas14} class containing $d_e$.
The event's peak cumulative rainfall across hexagons, $r_{\max}$, is then compared against the Harris-County-median NOAA Atlas 14 thresholds $\{\tau_{d^\ast, T}\}$ at the candidate return periods $T \in \{1, 2, 5, 10, 25, 50, 100, 200, 500, 1000\}$ years, and the event is assigned the largest $T$ for which the threshold is exceeded:
\begin{equation}\label{eq:rp_assign}
\mathrm{RP}_{\mathrm{event}} = \max\bigl\{\, T : r_{\max} \geq \tau_{d^\ast, T} \,\bigr\},
\end{equation}
with $\mathrm{RP}_{\mathrm{event}} = 0$ when no threshold is exceeded.

\subsection{Two-Stage Coreset Construction}\label{sec:4-coreset}

The choice of in-context examples is the central design decision of this paper.
A naive random sample of the simulation knowledge base ignores two properties of flood data: storm-magnitude imbalance leaves rare high-RP events sparse in any such subsample, and spatial autocorrelation among static features causes the picked hexagons to cluster geographically.
We address both with a two-stage construction: events are first dual-stratified by NOAA Atlas 14 return period and by most-affected watershed, the hydrologic unit of a storm's footprint.
Hexagons within each watershed are then sampled with a target-aware spatial selector that combines the simulated hydraulic response of each location with the static features describing its topographic and drainage controls.
The resulting coreset is the Cartesian product of $N_e$ events sampled in Stage 1 and $N_h$ hexagons selected per watershed in Stage 2, giving $N = N_e \times N_h$ rows per watershed.
Table~\ref{tab:ne_nh} lists the $(N_e, N_h)$ values used for each $N$.

\begin{table}[pos=!h]
\centering
\caption{Coreset decomposition: $(N_e, N_h)$ for each per-watershed size $N$.}
\label{tab:ne_nh}
\begin{tabular}{rrr}
\toprule
$N$ & $N_e$ & $N_h$ \\
\midrule
500    & 20 & 25   \\
1{,}000  & 25 & 40   \\
2{,}000  & 40 & 50   \\
5{,}000  & 50 & 100  \\
10{,}000 & 50 & 200  \\
50{,}000 & 50 & 1{,}000 \\
\bottomrule
\end{tabular}
\end{table}

The decomposition balances event diversity against per-watershed hexagon coverage within a context that fits all TFMs evaluated.
The largest size $N = 50\text{k}$ matches TabPFN-v2.5's native context limit, so every coreset-based model sees the same maximum context regardless of its own capacity ceiling.
Within this $N$ envelope, both $N_e$ and $N_h$ rise at small $N$ so the coreset captures both event types and watershed geometry.
$N_e$ saturates at $50$ once $N$ allows it, which keeps Stage 1 from depleting the rare-RP tail bins where the training pool itself is sparse.
$N_h$ then grows to keep $N = N_e \times N_h$ as $N$ continues to climb.

\subsubsection{Stage 1: Event Stratification}
Given $N_e$ events to select in Stage 1, we draw them from the training split, stratified jointly by RP bin and most-affected watershed.
Each watershed first gets a floor of $k_{\min} = \min(2, \lfloor N_e / n_{\mathrm{ws}} \rfloor)$ events, where $n_{\mathrm{ws}}$ is the number of watersheds.
The remaining slots are allocated across watersheds in proportion to their training-event counts (largest-remainder rounding), and within each watershed across its RP bins.
By construction, every watershed contributes at least $k_{\min}$ events, with within-watershed RP stratification preserving the pool's bin diversity.
Panel~(d) of Table~\ref{tab:exp1_method_sweep} shows that this stratification guarantees coverage at a small cost in mean accuracy.
Uniformly random draws of the same size reach slightly higher mean $R^2$, but they can leave watersheds and the rarest return-period bins unrepresented, and their accuracy on a far-OOD storm varies with the draw.

\subsubsection{Stage 2: Hexagon Selection}\label{sec:4-hs}
Within each watershed we evaluate five hexagon-selection methods: a Random baseline, Facility Location (FL), Spatial-Penalty FL (SP-FL), Depth-Stratified Sampling (Strat-Depth), and Depth-Augmented FL (FL-Depth).

\paragraph{Random.} Uniform random sampling, used as a baseline.

\paragraph{FL.} Greedy facility location on a feature-space RBF kernel \citep{lin2011submodular, wei2015submodularity, sener2018active}, evaluated over a candidate pool $\mathcal{H}_{\mathrm{pool}}$ of $3{,}000$ random hexagons subsampled from the watershed for tractability:
\begin{equation}\label{eq:fl1}
\mathcal{H}_{\mathrm{fl}} = \mathop{\arg\max}\limits_{|\mathcal{H}| = N_h} \sum_{h \in \mathcal{H}_{\mathrm{pool}}} \max_{s \in \mathcal{H}} K(x_h, x_s),
\end{equation}
\begin{equation}\label{eq:fl2}
K(x_h, x_s) = \exp\!\left(-\frac{\|x_h - x_s\|^2}{\sigma_f^2}\right),
\end{equation}
where $x_h$ is the z-scored static-feature vector of hexagon $h$ and $\sigma_f^2$ is set by the median heuristic on pairwise squared distances within the pool.

\paragraph{SP-FL.} Facility location on the same pool with a hard spatial-exclusion constraint:
\begin{equation}\label{eq:sp_fl}
\|p_{h_i} - p_{h_j}\|_{\mathrm{geo}} \geq r_{\min}, \qquad \forall\, h_i, h_j \in \mathcal{H}_{\mathrm{sp\_fl}},\ i \neq j,
\end{equation}
where $p_h$ is the hexagon centroid and $\|\cdot\|_{\mathrm{geo}}$ is haversine distance.
We fix $r_{\min} = 300$~m.
SP-FL can fall short of $N_h$ in small watersheds when the geometric constraint exhausts available hexagons (Hunting Bayou $56\%$ filled at $N = 50\text{k}$, Vince-Buffalo $71\%$).

\paragraph{Strat-Depth.} Stratified random sampling on a per-hexagon depth signal $y_h$.
The depth signal is the mean simulated peak depth at hexagon $h$ over the Stage 1 event set $\mathcal{E}$ for the current $N$:
\begin{equation}\label{eq:depth_signal}
y_h = \frac{1}{|\mathcal{E}|}\sum_{e \in \mathcal{E}} \mathrm{depth}_{h, e},
\end{equation}
where $\mathrm{depth}_{h, e}$ is the HEC-RAS-simulated peak depth at hexagon $h$ in event $e$.
Dry hexagons ($y_h \leq 0$) form a separate bin when present and wet hexagons are split into up to three quantile bins, for a total of at most four bins.
Selections are allocated equally per bin with largest-remainder allocation to the largest pools for the residual.

\paragraph{FL-Depth.} Facility location on the same pool with feature vector augmented by the z-scored depth signal from Eq.~\ref{eq:depth_signal}:
\begin{equation}\label{eq:fl_depth}
\tilde{x}_h = \bigl[\, x_h \,;\; z(y_h) \,\bigr], \qquad z(y) = \frac{y - \mu_y}{\sigma_y},
\end{equation}
where $\mu_y$ and $\sigma_y$ are the pool-wide mean and standard deviation of $y_h$.
The kernel bandwidth $\sigma_f^2$ is recomputed by the median heuristic on the augmented $\tilde{x}$ pool.

The depth signal in Strat-Depth and FL-Depth is computed over the candidate pool from the Stage 1 events, which come from the training split, so only training-split labels enter the selection and no depth from validation, test, or real-event queries is used.
This is target-aware sampling of the training set rather than leakage.
Random, FL, and SP-FL select on the eight static features alone and serve as the predictor-only comparison, so the gap between FL-Depth and FL in Section~\ref{sec:5} isolates the contribution of target awareness.

\subsection{Models}

\subsubsection{Reference Baselines}
Coreset-XGB is XGBoost \citep{chen2016xgboost} trained on a single coreset
of $N = 50\text{k}$ rows per watershed, with hyperparameters from
\citet{lee2024predicting}: $1{,}000$ histogram-based trees of maximum depth
$5$, learning rate $0.01$, column subsampling rate $0.3$, and $L_1$
regularization weight $10$.
Tuned gradient-boosted trees remain stronger than tabular deep-learning architectures on medium-sized to large tables with mostly numerical features \citep{grinsztajn2022why}, which is the regime of the $50\text{k}$-row, 11-feature coresets, so Coreset-XGB is a strong supervised baseline on the coreset.
Full-KB-XGB uses the same hyperparameters but
is trained on \emph{all} $350$ training events for that watershed, on
average ${\approx}\,6.9$~M rows.
We call it the supervised reference
rather than a strict ceiling, since several coreset-based models exceed
it on individual watersheds and on the far out-of-distribution storm.

\subsubsection{Vanilla TFMs}
We evaluate three vanilla TFMs.
TabPFN-v2.6 \citep{hollmann2025accurate,
priorlabs2025tabpfn26} is the latest release in the TabPFN family.
TabPFN-v2.5 \citep{grinsztajn2025tabpfn} provides a backbone version
contrast, and TabICL \citep{qu2025tabicl} extends the comparison to a
second TFM family.
At inference time the coreset $\mathcal{C}$ provides
the in-context examples:
\begin{equation}\label{eq:tfm}
\hat{y}_q = f_\theta\bigl(x_q;\, \{(x_i, y_i)\}_{i \in \mathcal{C}}\bigr)
\end{equation}
where $f_\theta$ is the pretrained transformer and $\theta$ is held fixed.
No gradient updates are applied at inference.

\subsubsection{Fine-Tuning Variants}
To test whether fine-tuning (FT) improves over vanilla inference, we add two FT
modes on both the TabPFN-v2.5 and TabPFN-v2.6 backbones.
TabPFN-FT-v2.5 / v2.6
applies per-watershed fine-tuning on each watershed's own FL-Depth
coreset.
TabPFN-FTLOO-v2.5 / v2.6 is fine-tuned across the other eight
watersheds, with each episode drawing context and query from one of those
eight sampled uniformly at random, so that on a held-out target both
context and weights are target-free.
Both modes share a single recipe:
$500$ episodes with fresh context and query batches of $4{,}000$ and
$2{,}048$ rows, AdamW at learning rate $10^{-5}$ and weight decay
$10^{-4}$, gradient clipping at $1.0$, bfloat16 autocast, and MSE loss on
context-normalized targets.

\subsection{Prediction Intervals}\label{sec:4-pi}
Point accuracy alone does not show whether a model knows when it is uncertain, so on the real storms we also report $90\%$ prediction intervals.
For Vanilla TabPFN-v2.6 the interval is the 5th to 95th percentile of the predictive distribution that the model returns for each query under the same FL-Depth context.
For a matched reference we retrain the Full-KB-XGB configuration as a quantile regressor with the pinball loss \citep{koenker1978regression} at the 5th, 50th, and 95th percentiles on the same training rows.
For a query set $\mathcal{Q}$ with lower and upper quantile predictions $\hat{q}_{0.05}(x)$ and $\hat{q}_{0.95}(x)$, the empirical coverage is
\begin{equation}\label{eq:coverage}
\mathrm{Cov}_{90} = \frac{1}{|\mathcal{Q}|} \sum_{q \in \mathcal{Q}} \mathbb{1}\bigl[\hat{q}_{0.05}(x_q) \leq y_q \leq \hat{q}_{0.95}(x_q)\bigr],
\end{equation}
and the mean width is the average of $\hat{q}_{0.95}(x_q) - \hat{q}_{0.05}(x_q)$ over $\mathcal{Q}$.
Both are reported overall and on the in-distribution and OOD slices.

%% file: 5_EXPERIMENTS_AND_RESULTS.tex
\section{Experiments and Results}\label{sec:5}
\subsection{Coreset Construction}
The coresets evaluated below come from the two-stage pipeline of Section~\ref{sec:4-coreset}, with event selection in Stage 1 and hexagon selection in Stage 2.

\begin{figure*}[pos=!h]
    \centering
    \includegraphics[width=0.8\linewidth]{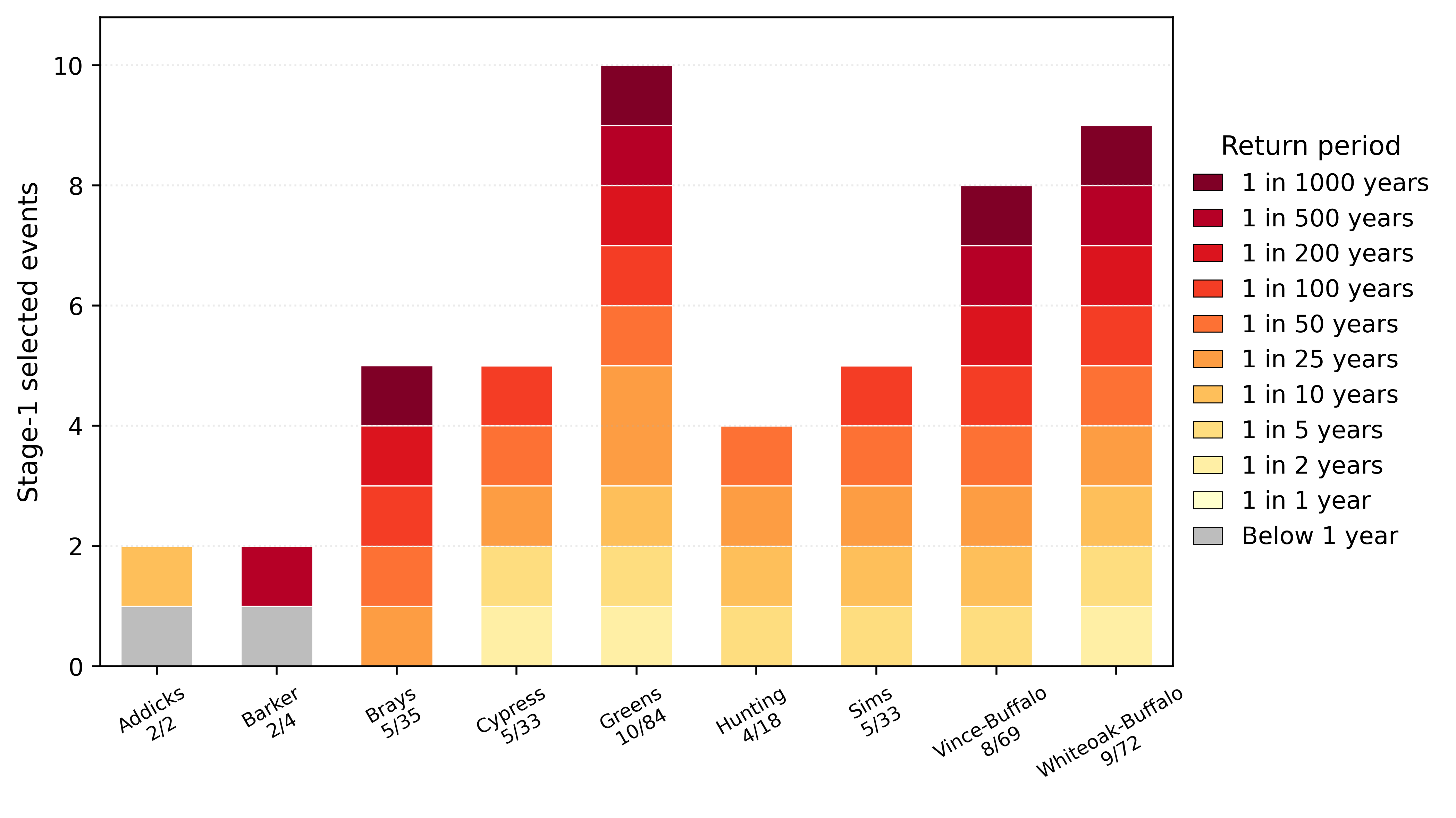}
    \caption{Per-watershed event allocation at $N_e = 50$. Bars stacked by return period bin, with x-axis labels showing the selected / training-pool ratio per watershed.}
    \label{fig:stage1}
\end{figure*}

Stage 1 selects $N_e = 50$ training events.
This is large enough for proportional allocation to give every watershed several events despite the uneven per-watershed training pools, and small enough for $N = N_e \times N_h$ to fit within TabPFN-v2.5's context cap.
The allocation in Figure~\ref{fig:stage1} combines proportional sampling with a floor of $k_{\min} = 2$, routing most of the sample to the larger bayou watersheds.
The floor catches Addicks and Barker, whose small training pools would otherwise round to zero and exclude these watersheds from the downstream protocols.
The below-1-year stratum visible at Addicks and Barker is a direct consequence of this floor on small pools.
Each pool spans only $2$ and $4$ RP bins respectively, including a below-1-year bin, so the floor forces both watersheds to draw bins they would otherwise miss.
The $500$-year and $1{,}000$-year classes each surface in four watersheds, providing rare-event coverage across the bayou network.

\begin{figure*}[pos=!h]
    \centering
    \includegraphics[width=\linewidth]{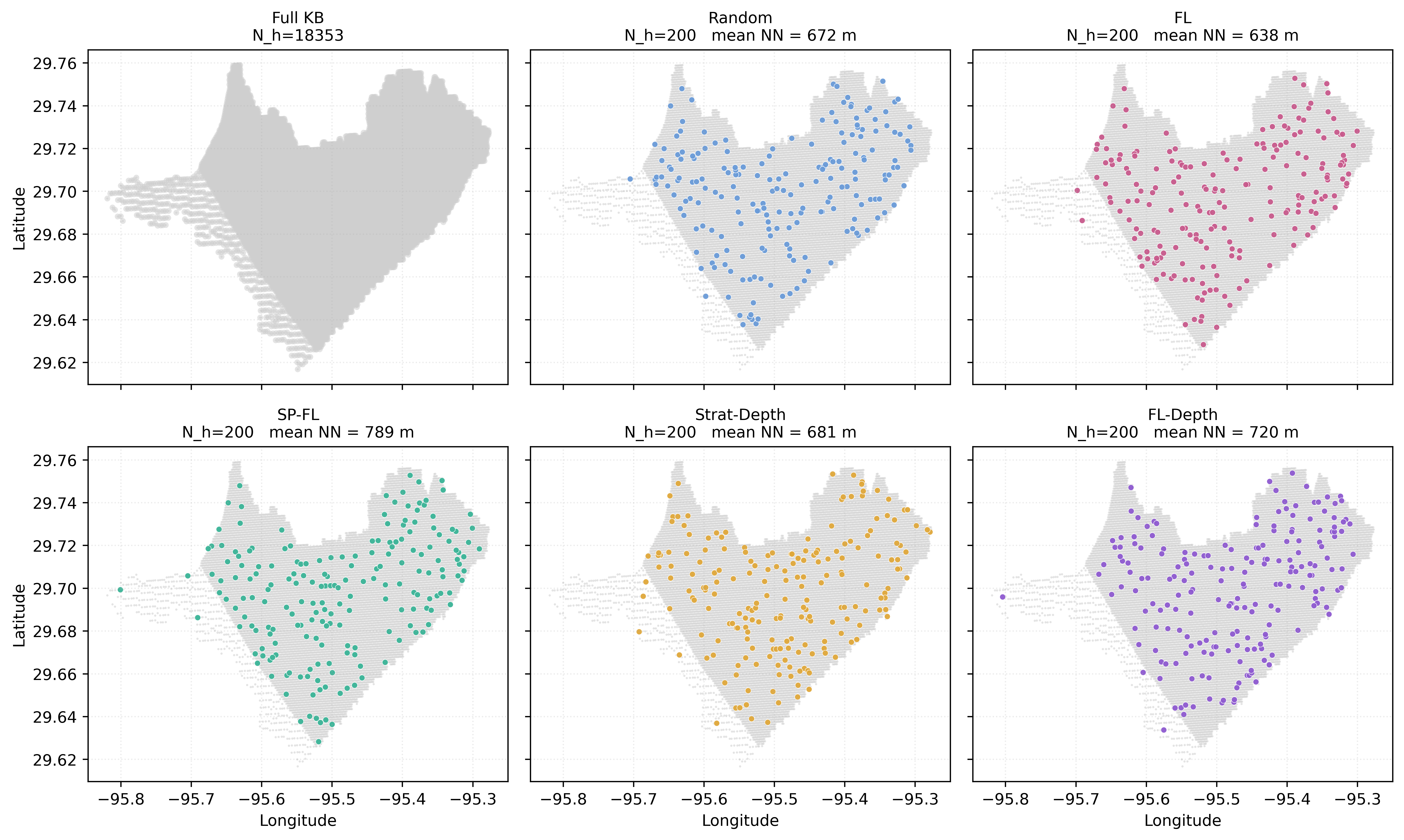}
    \caption{Spatial distribution of selected hexagons for Brays Bayou at $N = 10\text{k}$ ($N_h = 200$) under the five Stage 2 methods. Panel titles give the mean nearest-neighbor distance among the selected hexagons.}
    \label{fig:spatial_dist}
\end{figure*}

Stage 2 selects $N_h$ hexagons per watershed under five candidate selectors defined in Section~\ref{sec:4-hs}.
The selectors trade off feature-space coverage against spatial spread, illustrated for Brays Bayou at $N = 10\text{k}$ ($N_h = 200$) in Figure~\ref{fig:spatial_dist}.
FL clusters tightest because its feature-space objective does not penalize spatial proximity, so two nearby hexagons with distinct features can both be selected.
Random and Strat-Depth lack any spatial term and follow the underlying mesh density.
FL-Depth is more spread than these spatially neutral baselines because its feature vector includes the z-scored depth signal, which varies smoothly in space and therefore makes nearby hexagons redundant under the FL objective.
SP-FL has the widest spread under the $r_{\min} = 300$~m exclusion constraint.
The mean nearest-neighbor distance in each panel title quantifies this ordering, from $638$~m for FL to $789$~m for SP-FL.

\begin{table*}[pos=!h]
\centering
\caption{Mean $R^2$ across nine watersheds for the five hexagon selectors. Panels (a) and (b) sweep six coreset sizes $N$ on Vanilla TabPFN-v2.5 and Coreset-XGB, and panel (c) repeats the five selectors on the primary Vanilla TabPFN-v2.6 backbone at the $N = 50\text{k}$ operating point. Bold marks each column's best selector. FL-Depth has the highest cross-$N$ average for both swept models and the highest mean on v2.6, and XGB shows a smaller method spread than v2.5 at every $N$. Panel (d) keeps FL-Depth on the v2.6 backbone and replaces the stratified Stage 1 events by uniformly random draws of the same size.}
\label{tab:exp1_method_sweep}

\begin{subtable}{\linewidth}
\centering
\caption{Vanilla TabPFN-v2.5}
\label{tab:exp1_method_sweep_v25}
\begin{tabular}{lrrrrrrr}
\toprule
Method & $N=500$ & $N=1\text{k}$ & $N=2\text{k}$ & $N=5\text{k}$ & $N=10\text{k}$ & $N=50\text{k}$ & Mean \\
\midrule
Random      & 0.348 & 0.370 & 0.381 & \textbf{0.429} & 0.463 & 0.609 & 0.433 \\
FL          & 0.315 & 0.283 & 0.277 & 0.346 & 0.393 & 0.609 & 0.371 \\
SP-FL       & 0.320 & 0.307 & 0.298 & 0.377 & 0.417 & 0.602 & 0.387 \\
Strat-Depth & 0.381 & 0.442 & 0.453 & 0.409 & \textbf{0.471} & 0.611 & 0.461 \\
FL-Depth    & \textbf{0.455} & \textbf{0.475} & \textbf{0.469} & 0.402 & 0.418 & \textbf{0.621} & \textbf{0.473} \\
\bottomrule
\end{tabular}
\end{subtable}

\vspace{1em}

\begin{subtable}{\linewidth}
\centering
\caption{Coreset-XGB}
\label{tab:exp1_method_sweep_xgb}
\begin{tabular}{lrrrrrrr}
\toprule
Method & $N=500$ & $N=1\text{k}$ & $N=2\text{k}$ & $N=5\text{k}$ & $N=10\text{k}$ & $N=50\text{k}$ & Mean \\
\midrule
Random      & 0.400 & 0.427 & 0.425 & 0.486 & 0.524 & 0.601 & 0.477 \\
FL          & 0.367 & 0.376 & 0.384 & 0.474 & 0.515 & \textbf{0.609} & 0.454 \\
SP-FL       & 0.373 & 0.387 & 0.396 & 0.479 & 0.523 & 0.605 & 0.461 \\
Strat-Depth & 0.414 & 0.433 & 0.455 & \textbf{0.487} & \textbf{0.528} & 0.605 & 0.487 \\
FL-Depth    & \textbf{0.436} & \textbf{0.466} & \textbf{0.469} & 0.458 & 0.514 & 0.606 & \textbf{0.491} \\
\bottomrule
\end{tabular}
\end{subtable}

\vspace{1em}

\begin{subtable}{\linewidth}
\centering
\caption{Vanilla TabPFN-v2.6 at $N = 50\text{k}$.}
\label{tab:exp1_method_sweep_v26}
\begin{tabular}{lrrr}
\toprule
Method & Mean $R^2$ & Std & Best on \\
\midrule
Random      & 0.646 & 0.055 & 0 / 9 \\
FL          & 0.648 & 0.051 & 1 / 9 \\
SP-FL       & 0.639 & 0.051 & 0 / 9 \\
Strat-Depth & 0.649 & 0.051 & 1 / 9 \\
FL-Depth    & \textbf{0.663} & 0.059 & 7 / 9 \\
\bottomrule
\end{tabular}
\end{subtable}

\vspace{1em}

\begin{subtable}{\linewidth}
\centering\setlength{\tabcolsep}{4pt}
\caption{Stage 1 ablation on Vanilla TabPFN-v2.6 with FL-Depth hexagons. Stratified events (as in panels a to c) against uniformly random draws of the same $N_e$ training events (mean and standard deviation over four seeds). The last two columns are the real storms of Section~\ref{sec:5-exp3} at $N = 50\text{k}$.}
\label{tab:exp1_stage1_ablation}
\begin{tabular}{lrrrrrrrr}
\toprule
Stage 1 & $N=500$ & $N=1\text{k}$ & $N=2\text{k}$ & $N=5\text{k}$ & $N=10\text{k}$ & $N=50\text{k}$ & Harvey & Imelda \\
\midrule
Stratified & 0.465 & 0.509 & 0.527 & 0.531 & 0.581 & 0.663 & 0.558 & 0.430 \\
Random, mean & 0.493 & 0.523 & 0.535 & 0.552 & 0.589 & 0.669 & 0.537 & 0.479 \\
Random, std & 0.005 & 0.003 & 0.003 & 0.002 & 0.003 & 0.004 & 0.043 & 0.014 \\
\bottomrule
\end{tabular}
\end{subtable}
\end{table*}

To pick the downstream hexagon selector, we sweep Vanilla TabPFN-v2.5 as the representative TFM and Coreset-XGB as a tree-based control over all five methods at six $N$ values, reported in Table~\ref{tab:exp1_method_sweep}.
For v2.5 in panel~(a), FL-Depth wins at four of the six $N$ values and tops the cross-$N$ average at $R^2 = 0.473$, with Strat-Depth second and the pure facility-location methods FL and SP-FL well behind.
FL-Depth slips below Random at $N = 5\text{k}$ and below Strat-Depth at $N = 10\text{k}$, a mid-$N$ dip that appears on both backbones and closes again by $N = 50\text{k}$.
Outside that mid-$N$ window, depth-aware selectors dominate because the per-hexagon depth signal compresses each hexagon's response across the Stage 1 event mix into a single target-relevant dimension.
The Stage 1 events span below-1-year through $1{,}000$-year storms, so hexagons selected along this signal cover the flood-response surface directly.
Pure feature-space selection picks hexagons that look feature-diverse but can still cluster on flood response, leaving the target distribution under-covered.
The largest size $N = 50\text{k}$ is the operating point the rest of our experiments target.
It matches TabPFN-v2.5's native context cap and is the largest context that fits every TFM under parity, even though it represents only about $0.7\%$ of the per-watershed KB.
At this size every reasonable coreset spans the flood-response surface, the five methods converge into a narrow band, and FL-Depth still leads.
Coreset-XGB in panel~(b) follows the same broad ordering but with a much smaller method spread.
FL-Depth has a small early lead, but the methods converge quickly as $N$ grows because the tree ensemble absorbs row-selection noise into its fit rather than propagating it like an in-context TFM.
XGB therefore acts only as a control here, and we adopt FL-Depth as the default hexagon selector for all coreset-based models in the rest of the paper.
Panel~(c) reports the mean $R^2$, its cross-watershed standard deviation, and the number of watersheds won by each selector on the primary v2.6 backbone at $N = 50\text{k}$.
FL-Depth has the highest mean and wins seven of nine watersheds.
Paired Wilcoxon tests place it above each of the other four selectors ($p \leq 0.008$).
Random selection reaches $0.646$, so the backbone alone does not close the gap to the reference.

\begin{figure}[pos=!h]
    \centering
    \includegraphics[width=\linewidth]{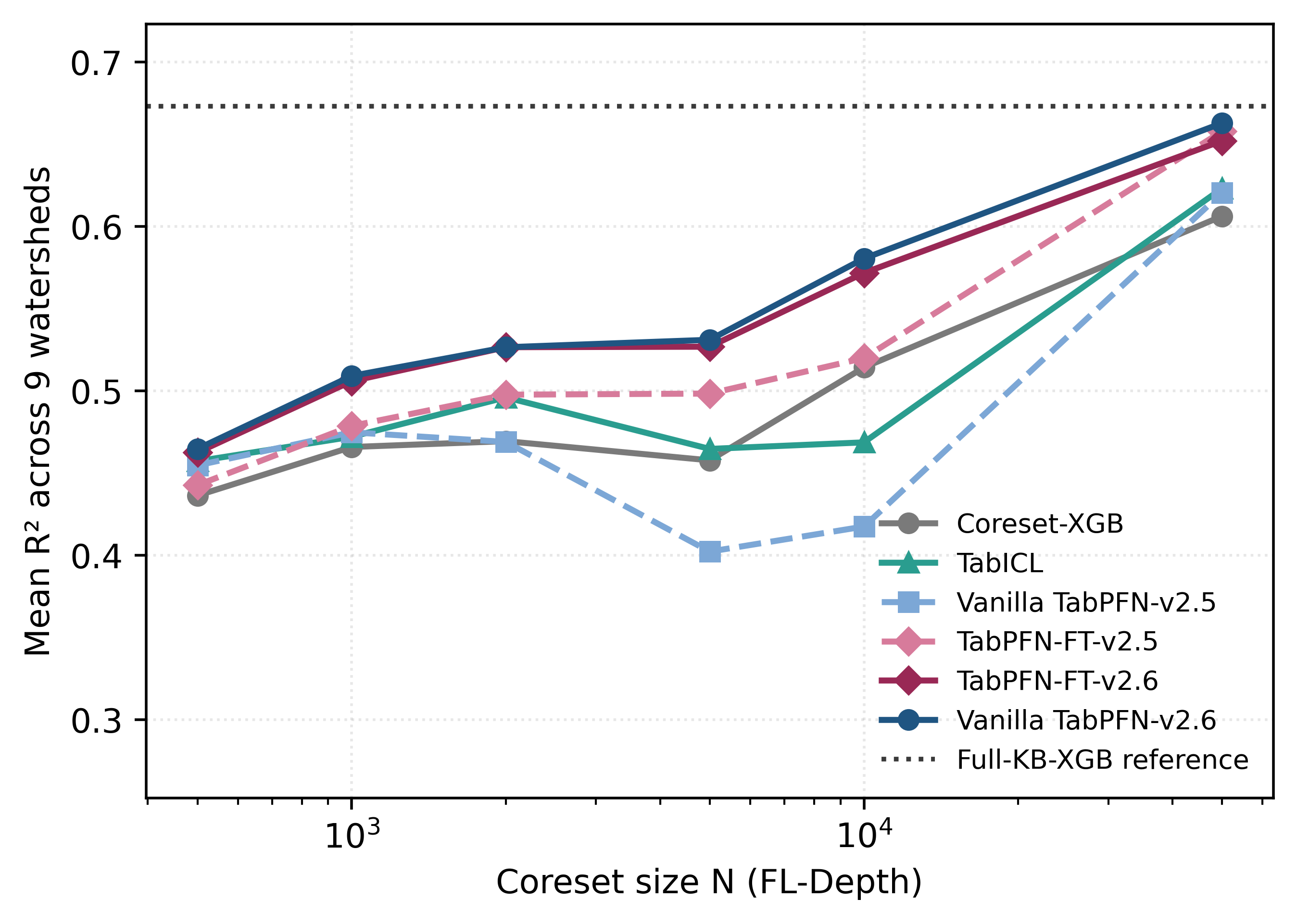}
    \caption{Within-watershed $R^2$ scaling with coreset size $N$. Mean $R^2$ across the nine watersheds for the six coreset-based models under FL-Depth, with the Full-KB-XGB reference shown as the dotted line.}
    \label{fig:exp1}
\end{figure}

Panels~(a) to~(c) together separate the contribution of the selector from that of the backbone.
Table~\ref{tab:exp1_method_sweep} varies the selector on a fixed backbone.
Experiment~1 below (Figure~\ref{fig:exp1}) varies the backbone on the fixed FL-Depth selector.
We measure each axis by its spread, the gap between the highest and lowest nine-watershed mean $R^2$ along that axis at a given $N$.
For $N \leq 2\text{k}$ the selector spread is larger ($0.14$ against $0.03$ at $N = 500$), so the choice of hexagons matters more when data are scarce.
For $N \geq 5\text{k}$ the order reverses.
At $N = 50\text{k}$ the selector spread stays below $0.025$ on every backbone while the backbone spread is $0.057$.
The headline $R^2$ is therefore set mainly by the backbone, and FL-Depth contributes mostly at small $N$.

Panel~(d) turns to the other half of the pipeline, the Stage 1 events.
It replaces the stratified event set with uniformly random draws of the same $N_e$ training events, over four seeds, at every $N$ and on both real storms.
Random draws score slightly higher in mean $R^2$ at every $N$ and on Imelda, so stratification buys no mean accuracy.
What it buys is coverage and determinism.
Every stratified set holds events centered on all nine watersheds, whereas a random draw can miss as many as four, and at $N = 500$ most draws contain no $1{,}000$-year event.
On the far-OOD Harvey the four random draws differ by almost $0.1$ in $R^2$, while the stratified set scores above their mean and has no seed to choose.
We keep stratification for these guarantees and read the small gap in panel~(d) as the price of reserving part of a $50$-event budget for the rare bins and the small watersheds.

\subsection{Experiment 1: Within-Watershed}\label{sec:5-exp1}

We investigate how close a coreset-based model gets to the watershed-level supervised reference and how fine-tuning shifts that gap.
Using the FL-Depth coreset across six values of $N$ from $500$ to $50\text{k}$, we evaluate three vanilla TFMs, two per-watershed fine-tuned TabPFN variants, and Coreset-XGB.
Full-KB-XGB sets the supervised reference, trained on all $350$ training events per watershed.
We trace how mean $R^2$ scales with $N$ to see whether any coreset-based model approaches the reference, then break the $N = 50\text{k}$ results down by watershed and quantify how much fine-tuning helps each TFM backbone.

\subsubsection{Aggregate Scaling with $N$}

The coreset-based models differ in how steeply they improve with $N$ and in which size brings them closest to the Full-KB-XGB reference.
Figure~\ref{fig:exp1} reports the within-watershed mean $R^2$ at each $N$ for the six coreset-based models and the reference.
Coreset-XGB rises monotonically with $N$ but plateaus well below the reference, since the tree ensemble has no pretrained prior and learns only from the coreset.
TabICL also rises with $N$ but stays below the TabPFN-v2.6 family across the sweep.
Its native context window is much larger than $50\text{k}$, so the parity cap we apply is conservative for TabICL and leaves it short of its optimum in this regime.
Vanilla TabPFN-v2.5 climbs through small $N$, underperforms in the intermediate $N$ range, then recovers at $N = 50\text{k}$, a v2.5 backbone weakness that the v2.6 backbone eliminates.
TabPFN-FT-v2.5 anchors weights to each target watershed and smooths the v2.5 mid-$N$ dip, but at $N = 50\text{k}$ it still trails Vanilla TabPFN-v2.6.
Vanilla TabPFN-v2.6 rises monotonically with $N$ and reaches $R^2 = 0.663$ at $N = 50\text{k}$, the highest among coreset-based models and statistically indistinguishable from the Full-KB-XGB reference at $0.673$.
The same fine-tuning recipe applied to v2.6 yields TabPFN-FT-v2.6, which tracks Vanilla TabPFN-v2.6 closely but lands slightly lower.
The v2.6 prior already encodes the per-watershed adjustment that fine-tuning extracts from v2.5, and additional fine-tuning over-fits the limited target signal in each watershed's coreset.
Two patterns dominate the sweep.
The backbone upgrade from v2.5 to v2.6 delivers a larger gain than fine-tuning v2.5 does, and fine-tuning v2.6 provides no benefit in this evaluation.
Vanilla TabPFN-v2.6 is therefore the strongest within-watershed model in this evaluation.
The stronger pretrained prior delivers more than per-watershed adaptation on the v2.5 backbone, and on the v2.6 backbone the prior is already too informative for fine-tuning to improve.

\subsubsection{Per-Watershed Model Comparison}\label{sec:5-exp1-perws}

At $N = 50\text{k}$ with the FL-Depth coreset, we evaluate per-watershed performance for each coreset-based model and the Full-KB-XGB reference.
Table~\ref{tab:exp1_perws} reports the $R^2$ values, and Table~\ref{tab:exp1_perws_err} summarizes $R^2$, RMSE, and MAE across watersheds, together with three measures restricted to the deepest cells.
By training-pool size the nine watersheds fall into three groups.
Addicks and Barker are small reservoirs with $\leq 4$ events each, Hunting is a small-pool bayou with $18$ events, and the other six bayous hold $33$ to $84$ events.
Volatility across watersheds tracks the TFM backbone.
Vanilla TabPFN-v2.5 has the widest cross-watershed standard deviation at $\sigma = 0.090$.
It reaches above the reference on Hunting and Barker where the supervised model is data-limited by small training pools, but falls well below it on Brays and Greens where larger pools give the reference more signal than v2.5's $50$-event coreset can extract.
TabICL is similarly volatile and drops below Coreset-XGB on the harder bayou watersheds, since its column-row attention is designed for much larger contexts than $50\text{k}$.
Per-watershed fine-tuning on the v2.5 backbone reduces $\sigma$ to $0.072$ by anchoring weights to each target watershed, and the v2.6 backbone reduces it to roughly $0.06$ regardless of fine-tuning.
Backbone strength rather than fine-tuning carries the stability gain.
Vanilla TabPFN-v2.6 wins six of the nine watersheds among coreset-based models, with TabPFN-FT-v2.5 taking Barker and Vince-Buffalo and Vanilla TabPFN-v2.5 taking Hunting.
More striking, Vanilla TabPFN-v2.6 exceeds the Full-KB-XGB reference on Barker, Hunting, and Sims while recovering $93\%$ to $100\%$ on the remaining six.
The three reference-beating watersheds share small-to-moderate training pools where the supervised reference itself is data-limited, leaving room for in-context inference with a strong prior to compensate.
The remaining six watersheds have larger pools that benefit Full-KB-XGB more than a fixed-size coreset can match.
At only $0.7\%$ of the per-watershed training pool, this makes the coreset approach data-efficient at the watershed level.
The $0.7\%$ is a count of rows, and rows repeat labels.
Hexagons in one mesh cell share that cell's depth label (Section~\ref{sec:3}), so the full knowledge base holds one distinct label for every $6.7$ rows on average.
The hexagons that FL-Depth selects mostly fall in different cells, so a coreset holds about four distinct labels for every five rows.
Counted in distinct labels, a $50\text{k}$ coreset covers about $37\%$ of a watershed's mesh cells on its $50$ training events, or about $5\%$ of the labels in the pool.
Scoring at the hexagon resolution, where shared labels are counted repeatedly, does not favor the TFM.
Re-scored at the mesh-cell level, with hexagon predictions averaged within each cell and compared with the shared label, both Vanilla TabPFN-v2.6 and Full-KB-XGB gain, and the TFM gains slightly more.
The paired Wilcoxon tests in the last row of Table~\ref{tab:exp1_perws} show that Vanilla TabPFN-v2.6 is statistically indistinguishable from the Full-KB-XGB reference ($p = 0.25$) and significantly better than Coreset-XGB and TabPFN-FT-v2.6.
Table~\ref{tab:exp1_perws_err} shows that the error metrics rank the models differently from $R^2$.
Every TFM has a lower MAE than the reference, so the typical error of the coreset models is smaller, while the reference keeps the lowest RMSE, so the largest errors of the coreset models are heavier.
TabPFN-FT-v2.5 has the lowest MAE of all models.
The last three columns of Table~\ref{tab:exp1_perws_err} look at the deepest cells, which matter most for damage.
The $R^2$ on cells deeper than $3$~ft leaves the ranking unchanged, with the reference slightly ahead of Vanilla TabPFN-v2.6.
On the deepest tenth of each watershed's cells every model under-predicts, with a mean bias error (MBE, prediction minus simulation) between $-2.0$ and $-4.5$~ft, and the two tree models under-predict most.
The critical success index $\mathrm{CSI} = H / (H + M + F)$ \citep{schaefer1990critical}, with hits $H$, misses $M$, and false alarms $F$ counted at the $1$~ft threshold, is higher for every TFM than for either tree model.

\begin{table*}[pos=!h]
\centering
\caption{Per-watershed $R^2$ at $N = 50\text{k}$ under FL-Depth, with the bottom rows giving the cross-watershed mean, standard deviation $\sigma$, and the exact two-sided Wilcoxon signed-rank $p$-value of each model against V-TabPFN-v2.6 over the nine paired watersheds, computed on unrounded values. Bold marks each row's coreset-based maximum, and italics mark cells where a coreset-based model exceeds the Full-KB-XGB reference. V-TabPFN denotes Vanilla TabPFN, and TabPFN-FT is the fine-tuned variant.}
\label{tab:exp1_perws}
\begin{tabular}{l*{7}{c}}
\toprule
Watershed & \makecell{Coreset\\XGB} & TabICL & \makecell{V-TabPFN\\v2.5} & \makecell{TabPFN-FT\\v2.5} & \makecell{TabPFN-FT\\v2.6} & \makecell{V-TabPFN\\v2.6} & \makecell{Full-KB\\XGB} \\
\midrule
Addicks          & 0.592 & 0.605 & 0.615 & 0.652 & 0.651 & \textbf{0.662} & 0.677 \\
Barker           & 0.670 & 0.727 & \textit{0.766} & \textbf{\textit{0.779}} & 0.727 & \textit{0.749} & 0.728 \\
Brays            & 0.557 & 0.526 & 0.508 & 0.586 & 0.578 & \textbf{0.596} & 0.617 \\
Cypress          & 0.631 & 0.617 & 0.589 & 0.642 & 0.655 & \textbf{0.675} & 0.697 \\
Greens           & 0.561 & 0.553 & 0.527 & 0.559 & 0.580 & \textbf{0.593} & 0.632 \\
Hunting          & 0.667 & \textit{0.757} & \textbf{\textit{0.765}} & \textit{0.762} & \textit{0.745} & \textit{0.748} & 0.737 \\
Sims             & 0.601 & 0.645 & 0.607 & \textit{0.668} & \textit{0.673} & \textbf{\textit{0.680}} & 0.661 \\
Vince-Buffalo    & 0.589 & 0.580 & 0.601 & \textbf{0.630} & 0.619 & 0.601 & 0.649 \\
Whiteoak-Buffalo & 0.589 & 0.600 & 0.608 & 0.643 & 0.642 & \textbf{0.662} & 0.663 \\
\midrule
Mean             & 0.606 & 0.623 & 0.621 & 0.658 & 0.652 & 0.663 & 0.673 \\
Std              & 0.041 & 0.076 & 0.090 & 0.072 & 0.057 & 0.059 & 0.041 \\
$p$ vs.\ V-TabPFN-v2.6 & 0.004 & 0.008 & 0.055 & 0.57 & 0.039 & -- & 0.25 \\
\bottomrule
\end{tabular}
\end{table*}

\begin{table*}[pos=!h]
\centering\small\setlength{\tabcolsep}{4pt}
\caption{Cross-watershed mean $\pm$ standard deviation of $R^2$, RMSE, and MAE at $N = 50\text{k}$ under FL-Depth, and of $R^2$, MBE, and CSI on the deepest cells. RMSE, MAE, and MBE are in feet, the native unit of the HEC-RAS depth output. Bold marks the best coreset-based model in each column, the smallest magnitude for MBE.}
\label{tab:exp1_perws_err}
\begin{tabular}{lcccccc}
\toprule
Model & $R^2$ & RMSE (ft) & MAE (ft) & $R^2$ ($>3$ ft) & MBE (top $10\%$) & CSI ($1$ ft) \\
\midrule
Coreset-XGB & 0.606 $\pm$ 0.041 & 2.711 $\pm$ 0.539 & 1.818 $\pm$ 0.363 & 0.406 $\pm$ 0.078 & $-4.252$ $\pm$ 1.045 & 0.900 $\pm$ 0.048 \\
TabICL & 0.623 $\pm$ 0.076 & 2.653 $\pm$ 0.623 & 1.471 $\pm$ 0.386 & 0.401 $\pm$ 0.118 & \textbf{$-2.020$ $\pm$ 0.617} & 0.934 $\pm$ 0.039 \\
Vanilla TabPFN-v2.5 & 0.621 $\pm$ 0.090 & 2.665 $\pm$ 0.673 & 1.442 $\pm$ 0.408 & 0.383 $\pm$ 0.125 & $-2.275$ $\pm$ 0.727 & 0.932 $\pm$ 0.042 \\
TabPFN-FT-v2.5 & 0.658 $\pm$ 0.072 & 2.532 $\pm$ 0.611 & \textbf{1.398 $\pm$ 0.375} & 0.426 $\pm$ 0.099 & $-2.850$ $\pm$ 0.676 & 0.935 $\pm$ 0.041 \\
TabPFN-FT-v2.6 & 0.652 $\pm$ 0.057 & 2.550 $\pm$ 0.563 & 1.457 $\pm$ 0.356 & 0.412 $\pm$ 0.102 & $-3.569$ $\pm$ 0.908 & 0.937 $\pm$ 0.040 \\
Vanilla TabPFN-v2.6 & \textbf{0.663 $\pm$ 0.059} & \textbf{2.512 $\pm$ 0.568} & 1.478 $\pm$ 0.357 & \textbf{0.449 $\pm$ 0.096} & $-3.276$ $\pm$ 0.992 & \textbf{0.938 $\pm$ 0.039} \\
Full-KB-XGB & 0.673 $\pm$ 0.041 & 2.472 $\pm$ 0.516 & 1.582 $\pm$ 0.340 & 0.459 $\pm$ 0.077 & $-4.470$ $\pm$ 0.994 & 0.914 $\pm$ 0.039 \\
\bottomrule
\end{tabular}
\end{table*}

\begin{table*}[pos=!hb]
\centering
\caption{Touching-watershed neighbors used by the \emph{geo} mode, derived from mesh-cell geometry.}
\label{tab:adjacency}
\begin{tabular}{lcl}
\toprule
Target watershed & $K_T$ & Neighbors \\
\midrule
Addicks Reservoir            & 3 & Barker, Cypress, Whiteoak-Buffalo \\
Barker Reservoir             & 4 & Addicks, Brays, Cypress, Whiteoak-Buffalo \\
Brays Bayou                  & 4 & Barker, Sims, Vince-Buffalo, Whiteoak-Buffalo \\
Cypress Creek                & 4 & Addicks, Barker, Greens, Whiteoak-Buffalo \\
Greens Bayou                 & 4 & Cypress, Hunting, Vince-Buffalo, Whiteoak-Buffalo \\
Hunting Bayou                & 3 & Greens, Vince-Buffalo, Whiteoak-Buffalo \\
Sims Bayou                   & 2 & Brays, Vince-Buffalo \\
Vince Bayou-Buffalo Bayou    & 5 & Brays, Greens, Hunting, Sims, Whiteoak-Buffalo \\
Whiteoak Bayou-Buffalo Bayou & 7 & Addicks, Barker, Brays, Cypress, Greens, Hunting, Vince-Buffalo \\
\bottomrule
\end{tabular}
\end{table*}

\subsection{Experiment 2: Cross-Watershed LOO}\label{sec:5-exp2}

We measure cross-watershed transfer by holding out one target watershed at a time.
For each held-out target $T$, the inference context of size $N$ is built from the other eight watersheds' $50\text{k}$-row random coresets, with two source-selection modes determining how those eight pools feed the context.
The \emph{geo} mode uses only the $K_T$ watersheds touching $T$ in the mesh-derived adjacency graph of Table~\ref{tab:adjacency}, each contributing $N / K_T$ rows.
The \emph{all} mode uses all eight other watersheds, each contributing $N / 8$ rows.
The \emph{geo} pool size $K_T$ varies by target, and the mode tests whether watersheds touching $T$ carry sufficiently representative hydrologic regimes that a smaller and more local pool matches the wider pool of all eight.
Hexagons within each contributing watershed are sampled at random, since Experiment 1 already characterizes selector effects.
Every event in a source coreset comes from the training split, and the target watershed is queried on the same $42$ test events as in Experiment 1 (Section~\ref{sec:3}).
Each synthetic storm covers the whole county, so a training event also produced depths in the target watershed, but those rows never enter the context.
The context and the query therefore share neither a location nor a storm.
We sweep eight context sizes from $1\text{k}$ to $50\text{k}$ and compare three model groups.
Vanilla TabPFN on both backbones consumes the cross-watershed context without parameter updates.
TabPFN-FTLOO on both backbones is fine-tuned across the other eight watersheds before inference, so that neither the target's data nor its weights enter the prediction.
Coreset-XGB is a tree-based baseline fitted on the same cross-watershed context.
The Full-KB-XGB reference from Experiment 1 is not reused here because each of its models was trained on its own watershed's full training rows and would leak target labels into a leave-one-out evaluation, and keeping every model at the same $50\text{k}$-row cross-watershed context isolates the transfer effect from data-volume differences.
The leave-one-out protocol breaks the spatial leakage that a random $k$-fold split would not catch \citep{roberts2017cv, meyer2018improving}, providing a notion of cross-watershed transfer principled with respect to the spatial structure of the data \citep{meyer2021predicting}.

\begin{figure*}[pos=!h]
    \centering
    \includegraphics[width=0.95\linewidth]{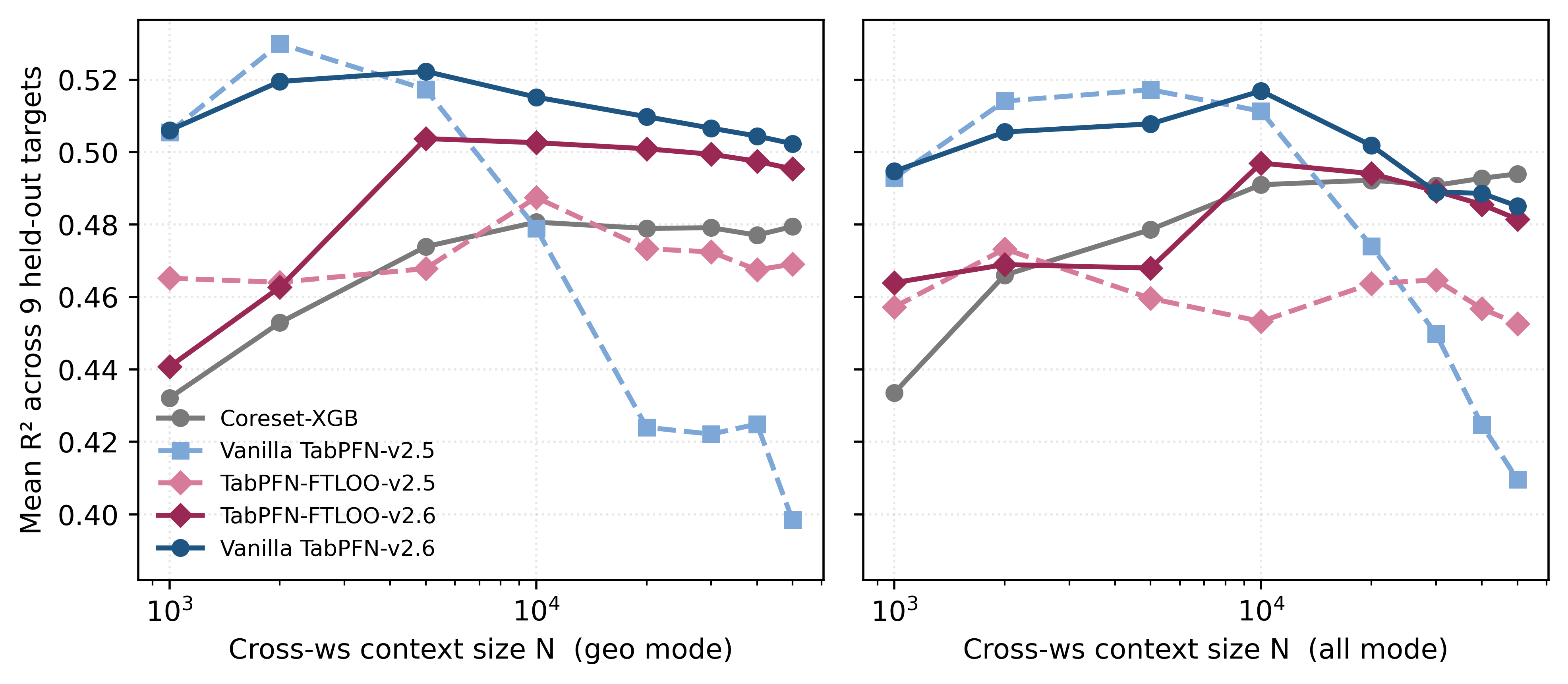}
    \caption{Cross-watershed LOO mean $R^2$ versus context size $N$ under the \emph{geo} (left) and \emph{all} (right) source-selection modes, averaged over nine held-out target watersheds.}
    \label{fig:exp2}
\end{figure*}

\begin{table*}[pos=!h]
\centering
\caption{Cross-watershed LOO $R^2$ at $N \in \{2\text{k}, 10\text{k}, 50\text{k}\}$, averaged over nine held-out target watersheds under the \emph{geo} and \emph{all} source-selection modes. Bold marks each row's maximum.}
\label{tab:exp2_smalllarge}
\begin{tabular}{llccccc}
\toprule
Mode & $N$ & \makecell{Coreset\\XGB} & \makecell{V-TabPFN\\v2.5} & \makecell{TabPFN-FTLOO\\v2.5} & \makecell{TabPFN-FTLOO\\v2.6} & \makecell{V-TabPFN\\v2.6} \\
\midrule
\multirow{3}{*}{\emph{geo}} & $2\text{k}$  & 0.4529 & \textbf{0.5300} & 0.4641 & 0.4627 & 0.5195 \\
                            & $10\text{k}$ & 0.4807 & 0.4790 & 0.4875 & 0.5026 & \textbf{0.5152} \\
                            & $50\text{k}$ & 0.4795 & 0.3984 & 0.4691 & 0.4954 & \textbf{0.5023} \\
\midrule
\multirow{3}{*}{\emph{all}} & $2\text{k}$  & 0.4660 & \textbf{0.5141} & 0.4733 & 0.4690 & 0.5056 \\
                            & $10\text{k}$ & 0.4910 & 0.5113 & 0.4533 & 0.4970 & \textbf{0.5169} \\
                            & $50\text{k}$ & \textbf{0.4940} & 0.4097 & 0.4527 & 0.4815 & 0.4851 \\
\bottomrule
\end{tabular}
\end{table*}

Cross-watershed performance separates cleanly by backbone.
Vanilla TabPFN-v2.5 peaks at the smallest $N$ and decays steadily as the context grows, while Vanilla TabPFN-v2.6 holds $R^2$ between $0.50$ and $0.52$ across the entire sweep and takes the lead from $N = 10\text{k}$ onward.
Figure~\ref{fig:exp2} traces the full sweep and Table~\ref{tab:exp2_smalllarge} pins three landmark $N$ values under both modes.
The TabPFN-FTLOO variants track their backbones without overtaking Vanilla TabPFN-v2.6, so leave-one-out fine-tuning offers no usable lift on the stronger backbone, mirroring the within-watershed FT finding from Experiment 1.
The mode gap stays narrow across the sweep.
The one exception is Coreset-XGB at $N = 50\text{k}$ under \emph{all}, where the wider and more diverse training pool benefits tree fitting enough to overtake every TFM.
Outside that corner, the narrow mode gap supports the spatial-locality intuition behind \emph{geo}'s design.
Touching watersheds carry enough representative variation to substitute for the full eight-watershed pool in cross-watershed transfer.
Vanilla TabPFN-v2.6 is therefore the strongest cross-watershed model, transferring leak-free to a held-out target within the same metropolitan region at the same $50\text{k}$-row context size as within-watershed inference.
As in Experiment 1, per-task fine-tuning brought no benefit on the stronger backbone in this evaluation.

\subsection{Experiment 3: Real Events}\label{sec:5-exp3}

We evaluate five coreset-based models and the Full-KB-XGB reference on Hurricane Harvey and Tropical Storm Imelda, with the results summarized in Table~\ref{tab:exp3}.
Each TFM uses its own watershed's coreset as in-context examples.
Under the three-variable envelope defined in Section~\ref{sec:3}, Harvey's storm profile lies entirely outside the synthetic training range and therefore serves as a fully out-of-distribution test.
Imelda has cells both inside and outside the envelope across the nine watersheds.

\begin{table}[pos=!h]
\centering
\caption{Real-event $R^2$ on Hurricane Harvey and Tropical Storm Imelda, averaged over the nine watersheds. Two TFM variants exceed the Full-KB-XGB reference on Harvey, while the reference reclaims the lead on Imelda.}
\label{tab:exp3}
\begin{tabular}{lrr}
\toprule
Model & Harvey & Imelda \\
\midrule
Coreset-XGB           & 0.471 & 0.430 \\
TabICL                & 0.481 & 0.380 \\
Vanilla TabPFN-v2.5   & 0.487 & 0.404 \\
Vanilla TabPFN-v2.6   & 0.558 & 0.430 \\
TabPFN-FT-v2.5        & 0.570 & 0.457 \\
Full-KB-XGB           & 0.503 & 0.528 \\
\bottomrule
\end{tabular}
\end{table}

On Harvey the two top TFM models exceed the Full-KB-XGB reference.
The TabPFN pretrained prior generalizes more reliably than tree-based fitting under sharp distribution shift, since XGB must extrapolate beyond its training envelope while the TFM stays inside its own prior.
The backbone upgrade from v2.5 to v2.6 delivers $+0.071$ $R^2$ on Harvey, most of the $+0.083$ that fine-tuning adds to the v2.5 backbone, which leaves TabPFN-FT-v2.5 only $0.012$ ahead of Vanilla TabPFN-v2.6.
On Imelda the Full-KB-XGB reference reclaims the lead.
The three top models converge within $0.02$ $R^2$ on the in-distribution slice, while on the OOD slice the reference leads Vanilla TabPFN-v2.6 by $+0.13$ $R^2$ in every watershed.
Imelda is a milder departure from the synthetic envelope than Harvey, since about half of its cells fall inside a 95\% Mahalanobis ellipsoid of the training pool while none of Harvey's do, so its OOD cells remain close enough for the tree model fitted on $6.9$M rows to interpolate.
The TFM advantage therefore appears under far-OOD shift rather than moderate shift.
TabPFN-FT-v2.5 keeps a small edge over Vanilla TabPFN-v2.6 on both storms, the only place in the evaluation where fine-tuning the v2.5 backbone still shows measurable value.
The two storms thus partition the real-event regime by the severity of distribution shift.
Coreset TFMs hold the advantage under far-OOD conditions, while tree-based models with full training access remain hard to beat when the shift is moderate or absent.

\begin{figure*}[pos=!h]
    \centering
    \includegraphics[width=\linewidth]{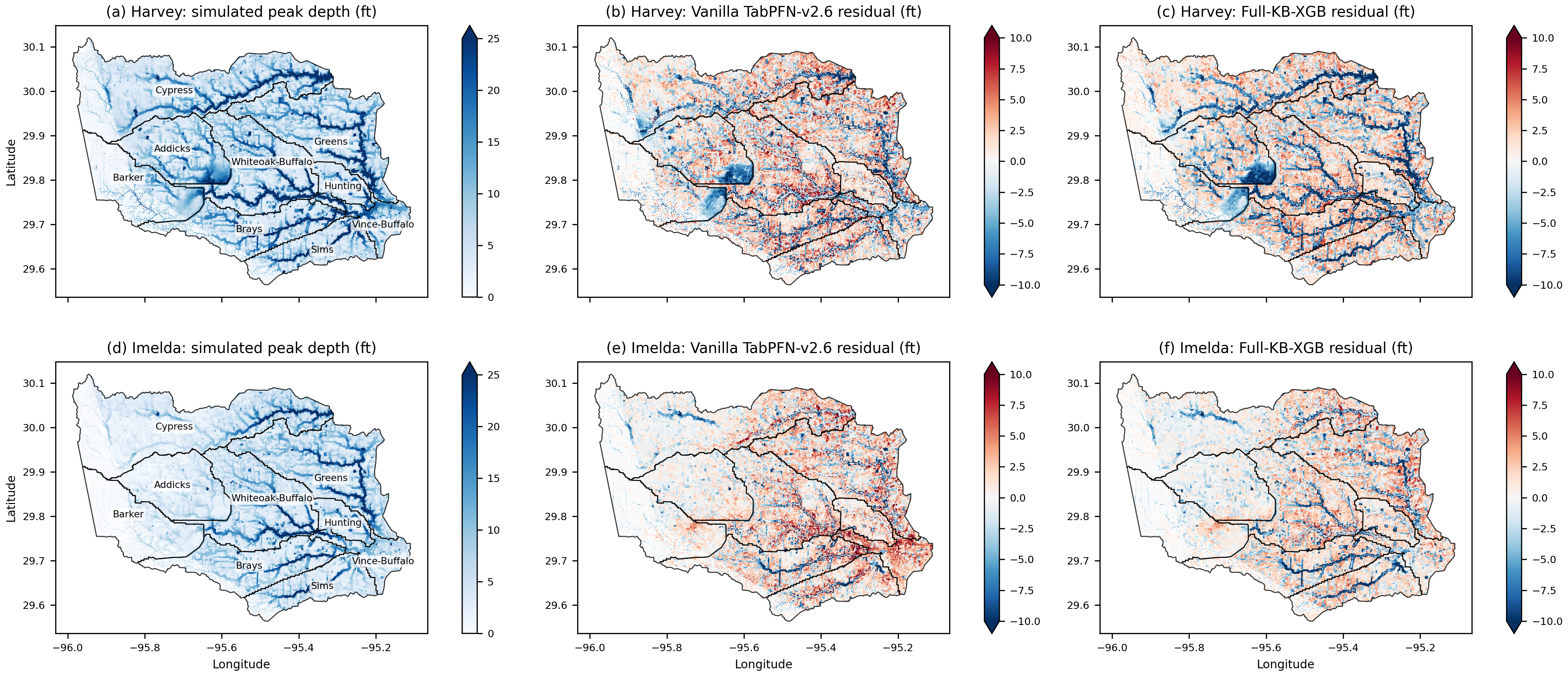}
    \caption{Simulated peak depth (a, d) and residual (prediction minus HEC-RAS depth, ft) of Vanilla TabPFN-v2.6 (b, e) and Full-KB-XGB (c, f) on Hurricane Harvey (top) and Tropical Storm Imelda (bottom) at $N = 50\text{k}$ under FL-Depth. All residual panels share one symmetric color scale, and black lines mark the watershed boundaries of Figure~\ref{fig:watersheds}.}
    \label{fig:residual_map}
\end{figure*}

The storm-level $R^2$ values say how far the two models differ on each storm, but not where.
Figure~\ref{fig:residual_map} maps the residual of the two leading models on both storms.
The channels carry the largest errors for both models on both storms, consistent with the under-prediction of the deepest cells in Table~\ref{tab:exp1_perws_err}.
The two models differ on the overland cells.
On Harvey the reference under-predicts in every watershed, while the TFM residual changes sign from one watershed to the next.
On Imelda the pattern reverses.
The TFM over-predicts in every watershed, by up to $3$~ft on average in Vince-Buffalo and most strongly over the overland cells, while the reference is nearly unbiased.
The moderate-shift deficit is therefore a systematic positive bias concentrated in the south-eastern watersheds rather than scatter.
The hybrid listed in Section~\ref{sec:6}, a tree model fitted to the TFM residual on in-distribution cells, is designed to remove a bias of this kind.

A systematic bias of this size also bears on the calibration of the prediction intervals.
Table~\ref{tab:intervals} reports the coverage and mean width of the $90\%$ prediction intervals of Section~\ref{sec:4-pi}.
On Harvey neither model reaches nominal coverage, so far-OOD shift degrades the calibration of both, while on Imelda the quantile reference is better calibrated than the TFM at similar interval width, on eight of nine watersheds for both storms.
The TFM intervals widen on OOD cells but remain somewhat over-confident relative to a supervised quantile model with full training access.

\begin{table*}[pos=!h]
\centering
\caption{Empirical coverage and mean width (ft) of $90\%$ prediction intervals on the real storms, averaged over the nine watersheds. Harvey has no in-distribution cells.}
\label{tab:intervals}
\begin{tabular}{llcccc}
\toprule
 & & \multicolumn{2}{c}{Vanilla TabPFN-v2.6} & \multicolumn{2}{c}{Quantile Full-KB-XGB} \\
\cmidrule(lr){3-4} \cmidrule(lr){5-6}
Event & Slice & Coverage & Width & Coverage & Width \\
\midrule
Harvey & overall (OOD)  & 0.833 & 11.4 & 0.862 & 12.4 \\
Imelda & overall        & 0.841 & 9.4  & 0.889 & 9.4  \\
Imelda & in-distribution & 0.844 & 6.9  & 0.851 & 6.8  \\
Imelda & OOD            & 0.841 & 10.1 & 0.898 & 10.0 \\
\bottomrule
\end{tabular}
\end{table*}

All runs use one NVIDIA RTX A6000 (48~GB) with a 64-core AMD EPYC 7702P and 1~TB of RAM (Python 3.10, PyTorch 2.5.1, XGBoost 3.0.2, TabPFN 7.1.1, TabICL 2.1.0), and every prediction time covers the same $827{,}540$ query rows.
On computational cost the tree models win outright.
Full-KB-XGB trains and predicts in seconds, the TFMs take one to three minutes per watershed at $23$ to $30$~GB of GPU memory, and fine-tuning adds two to three minutes more without improving accuracy.
The claim of this paper is data efficiency, the number of training rows a model needs to reach a given accuracy.
On that measure, Table~\ref{tab:cost} shows that Vanilla TabPFN-v2.6 reaches a mean $R^2$ of $0.663$ from $50\text{k}$ context rows, statistically indistinguishable from the $0.673$ that Full-KB-XGB obtains from $6.9$M training rows, while Coreset-XGB on the same $50\text{k}$ rows reaches only $0.606$.
At equal data the pretrained prior is therefore worth about $0.06$ in $R^2$, and at equal accuracy the TFM needs less than $1\%$ of the rows, at a total cost of about two minutes per watershed.
The coreset TFM is the choice when a new watershed must be served from a few dozen simulated events, and the full-data tree model when the complete simulation archive already exists and latency matters.

\begin{table*}[pos=!h]
\centering
\caption{Per-watershed computational cost at $N = 50\text{k}$, averaged over the nine watersheds. Mean $R^2$ is the nine-watershed average from Table~\ref{tab:exp1_perws}. Fit time is model training for XGB and context loading for the TFMs. Fine-tuning time is the one-time cost of the $500$-episode adaptation. Prediction time covers the same $827{,}540$ query rows for every model. GPU memory was not logged for the XGB models.}
\label{tab:cost}
\begin{tabular}{lrrrrrr}
\toprule
Model & \makecell{Train / context\\rows} & \makecell{Mean\\$R^2$} & \makecell{Fine-tune\\(s)} & \makecell{Fit\\(s)} & \makecell{Predict\\(s)} & \makecell{Peak GPU\\(GB)} \\
\midrule
Coreset-XGB         & 50k  & 0.606 & --  & 1.2 & 0.2   & --   \\
TabICL              & 50k  & 0.623 & --  & 1.0 & 163.7 & 29.5 \\
Vanilla TabPFN-v2.5 & 50k  & 0.621 & --  & 0.6 & 87.2  & 23.1 \\
TabPFN-FT-v2.5      & 50k  & 0.658 & 120 & 0.6 & 87.1  & 23.1 \\
TabPFN-FT-v2.6      & 50k  & 0.652 & 191 & 0.9 & 120.4 & 23.1 \\
Vanilla TabPFN-v2.6 & 50k  & 0.663 & --  & 0.9 & 120.4 & 23.1 \\
Full-KB-XGB         & 6.9M & 0.673 & --  & 9.3 & 0.1   & --   \\
\bottomrule
\end{tabular}
\end{table*}

%% file: 6_CONCLUSION.tex
\section{Conclusion}\label{sec:6}

We present a domain-aware coreset construction pipeline that conditions a tabular foundation model on a small fraction of the training rows used by a watershed-level supervised baseline.
With a coreset of about $0.7\%$ of the per-watershed training pool, Vanilla TabPFN-v2.6 reaches a mean $R^2$ of $0.663$ across nine Houston-area watersheds, statistically indistinguishable from the $0.673$ of a Full-KB-XGB reference trained on the full pool.
This makes Vanilla TabPFN-v2.6 the optimal TFM in our evaluation.
Three design elements shape this result, with different roles.
A sufficiently strong pretrained backbone sets the accuracy at the operating point, and in this evaluation per-task fine-tuning of that backbone provided no measurable benefit.
Target-aware spatial selection via FL-Depth picks hexagons that span the flood-response surface, and matters most at small context sizes.
Dual stratification of storm events by NOAA Atlas 14 return period and most-affected watershed does not raise mean accuracy, since uniformly random event draws of the same size score slightly higher.
It guarantees instead that every watershed and nearly every return-period bin is represented in the context, and it removes the seed dependence that random draws show on a far-OOD storm.
The same model also transfers to a held-out watershed within the same metropolitan region by drawing in-context examples from its touching neighbors, without any gradient updates.

The three experiments map the operating range of the candidates.
In the within-watershed evaluation, all coreset TFMs improve with context size $N$, but the trajectory depends on the backbone.
The TabPFN-v2.6 family rises monotonically across the sweep, while the TabPFN-v2.5 family dips through the intermediate $N$ range and recovers only at $N = 50\text{k}$.
Vanilla TabPFN-v2.6 reaches the highest mean $R^2$ among coreset models, statistically indistinguishable from the supervised reference, and per-watershed fine-tuning provided no further benefit in this evaluation.
In the cross-watershed leave-one-out evaluation, Vanilla TabPFN-v2.5 peaks at the smallest $N$ and decays as the context grows, while Vanilla TabPFN-v2.6 stays stable between $0.50$ and $0.52$ across the full sweep and leads the coreset-trained supervised baseline at most context sizes in both source-selection modes.
LOO fine-tuning likewise provided no benefit here.
Two real storms confirm the pattern, with TFMs exceeding the supervised reference on far-OOD Hurricane Harvey and the reference reclaiming the lead on Tropical Storm Imelda, which mixes in-distribution cells with cells only moderately outside the synthetic envelope.
Coreset TFMs therefore hold an advantage under far-OOD conditions, while supervised models with full training access remain hard to beat when the shift is moderate or absent.

Several boundaries of the current scope motivate future work.
Vanilla TabPFN-v2.6 has higher cross-watershed variance than the Full-KB-XGB reference ($\mathrm{std} = 0.059$ versus $0.041$), so its higher mean comes with a wider per-watershed worst case.
The $50\text{k}$-row context cap is well matched to per-watershed inference but leaves room to explore the larger native windows of TabPFN-v2.6 at roughly $100\text{k}$ and TabICL at higher still.
The real-event component covers two storms that span the OOD and in-distribution regimes, and the cross-watershed evaluation draws on nine hydrologically distinct but climatically similar watersheds in one metropolitan region, so the transfer results are intra-regional. Future work includes evaluating on a wider set of historical storms from this and other regions, testing transfer across distinct hydroclimatic regions to probe the broader transfer-learning capability of coreset TFMs, and exploring hybrids that combine TFM in-context inference with tree-based fitting on in-distribution residuals.
The $90\%$ prediction intervals of Section~\ref{sec:5-exp3} under-cover on the far-OOD storm for both model classes and are somewhat over-confident for the TFM on the mixed storm, so calibrating the native TFM distribution, for example by conformal adjustment on a held-out event set, is a direct next step.
Area of Applicability analysis \citep{meyer2021predicting} would extend the OOD characterization to settings where real events arrive on a different grid, and a broader set of tabular deep-learning baselines would further place the framework.